\documentclass[sigconf]{acmart}
\usepackage{tikz}
\usepackage{booktabs}
\usepackage{algorithm}
\usepackage{algpseudocode}
\usepackage{amsmath}
\usepackage{caption}
\usetikzlibrary{shapes.geometric, arrows, positioning, shapes.multipart}

\AtBeginDocument{%
  \providecommand\BibTeX{{%
    \normalfont B\kern-0.5em{\scshape i\kern-0.25em b}\kern-0.8em\TeX}}}

\copyrightyear{2024}
\acmYear{2024}
\setcopyright{rightsretained}
\acmConference[CODS-COMAD Dec '24]{8th International Conference on Data Science and Management of Data (12th ACM IKDD CODS and 30th COMAD)}{December 18--21, 2024}{Jodhpur, India}
\acmBooktitle{8th International Conference on Data Science and Management of Data (12th ACM IKDD CODS and 30th COMAD) (CODS-COMAD Dec '24), December 18--21, 2024, Jodhpur, India}
\acmPrice{}
\acmDOI{10.1145/3703323.3703743}
\acmISBN{979-8-4007-1124-4/24/12}

\begin{document}

\title{Optimizing Fantasy Sports Team Selection with Deep Reinforcement Learning}

\author{Shamik Bhattacharjee}
\authornote{Both authors contributed equally to this research.}
\affiliation{%
  \institution{Dream11}
  \city{Mumbai}
  \state{Maharashtra}
  \country{India}}
\email{shamik.bhattacharjee@dream11.com}

\author{Kamlesh Marathe}
\authornotemark[1]
\affiliation{%
  \institution{Dream11}
  \city{Mumbai}
  \state{Maharashtra}
  \country{India}}
\email{kamlesh.marathe@dream11.com}

\author{Nilesh Patil}
\affiliation{%
  \institution{Dream11}
  \city{Mumbai}
  \state{Maharashtra}
  \country{India}}
\email{nilesh.patil@dream11.com}

\author{Hitesh Kapoor}
\affiliation{%
  \institution{Dream11}
  \city{Mumbai}
  \state{Maharashtra}
  \country{India}}
\email{hitesh.kapoor@dream11.com}

\begin{abstract}
Fantasy sports, particularly fantasy cricket, have garnered immense popularity in India in recent years, offering enthusiasts the opportunity to engage in strategic team-building and compete based on the real-world performance of professional athletes. In this paper, we address the challenge of optimizing fantasy cricket team selection using reinforcement learning (RL) techniques. By framing the team creation process as a sequential decision-making problem, we aim to develop a model that can adaptively select players to maximize the team's potential performance. Our approach leverages historical player data to train RL algorithms, which then predict future performance and optimize team composition. This not only represents a huge business opportunity by enabling more accurate predictions of high-performing teams but also enhances the overall user experience. Through empirical evaluation and comparison with traditional fantasy team drafting methods, we demonstrate the effectiveness of RL in constructing competitive fantasy teams. Our results show that RL-based strategies provide valuable insights into player selection in fantasy sports.
\end{abstract}





\maketitle

\section{Introduction}
Cricket is one of the most popular sports globally, boasting a fan base exceeding a billion individuals. The Indian sub-continent alone constitutes the vast majority of these fans, underscoring the region's deep-rooted passion for the sport \cite{popularity}. This immense popularity has fueled a significant rise in the fantasy sports domain within India, particularly highlighted by the success of our company. Our company, a leading fantasy sports platform, boasts a user base exceeding 200 million, illustrating the substantial online engagement where fans create and manage their dream teams.

India is currently the fastest-growing fantasy sports market worldwide, with projections indicating a compound annual growth rate (CAGR) of 33\%, reaching 500 million users by FY27 \cite{72b}. Within this burgeoning market, cricket stands out as the leading sport, capturing the largest proportion of registered users. In 2023, an estimated 130 million users participated in fantasy cricket, accounting for 85\% of the total fantasy sports user base in India ~\cite{india_perc}. This dramatic growth highlights the transformative impact of fantasy sports on the cricket economy in India, creating significant business opportunities and paving the way for new innovations in this domain.

Fantasy sports is a type of online game where participants create virtual teams composed of real-life athletes from professional sports leagues. Participants compete against each other by joining contests, which can range from head-to-head matchups with two users to large-scale contests involving several million users. The success of a participant's fantasy team depends on the performance of the selected players in actual sporting events.

Participants begin by creating their fantasy teams, selecting players from the actual teams competing in a particular sporting event. These selections are made based on the expected performance of the players, while adhering to a budget constraint and/or following player-type-based restrictions. Once the fantasy teams are created, participants enter contests, which are structured competitions in which users play against one another. Contests vary in size, with smaller contests hosting as few as two participants and larger contests accommodating millions of users. During the sporting event, each player's performance is tracked, and fantasy points are awarded based on their contributions on the field. Points can be earned from various actions such as runs scored, wickets taken, and other sport-specific achievements. The user with the highest total fantasy points at the end of the contest emerges as the winner. Contest rankings determine the distribution of prize amounts among participants. Winners of contests receive prize money, which is typically distributed according to the final rankings. Larger contests may offer higher prize pools, attracting more participants and increasing the level of competition.

Overall, fantasy sports combine elements of sports knowledge, statistical analysis, and strategic decision-making. Participants use their knowledge of players and teams to construct fantasy teams that maximize potential points within the constraints set by the game's rules. The outcome of a contest hinges on the performance of the chosen players in the real sporting event, creating a dynamic and engaging experience for users. This form of interactive gaming has become increasingly popular worldwide, offering opportunities for sports enthusiasts to test their knowledge and compete against others in a virtual environment.

As mentioned, the task of creating a fantasy team demands deep analysis and a thorough understanding of various elements such as player performance histories, tournament dynamics, venue specifics, and pitch conditions. Research indicates that the formulation of a competitive fantasy squad is far from arbitrary. Empirical data suggests a stark underperformance of teams assembled randomly compared to those crafted with strategic insight by seasoned users, highlighting the complex skills needed to do well in fantasy sports leagues. This particularly affects new users or those with less up-to-date knowledge of player or match/tournament statistics.

\begin{figure}[h]
    \centering
    \includegraphics[width=1\linewidth]{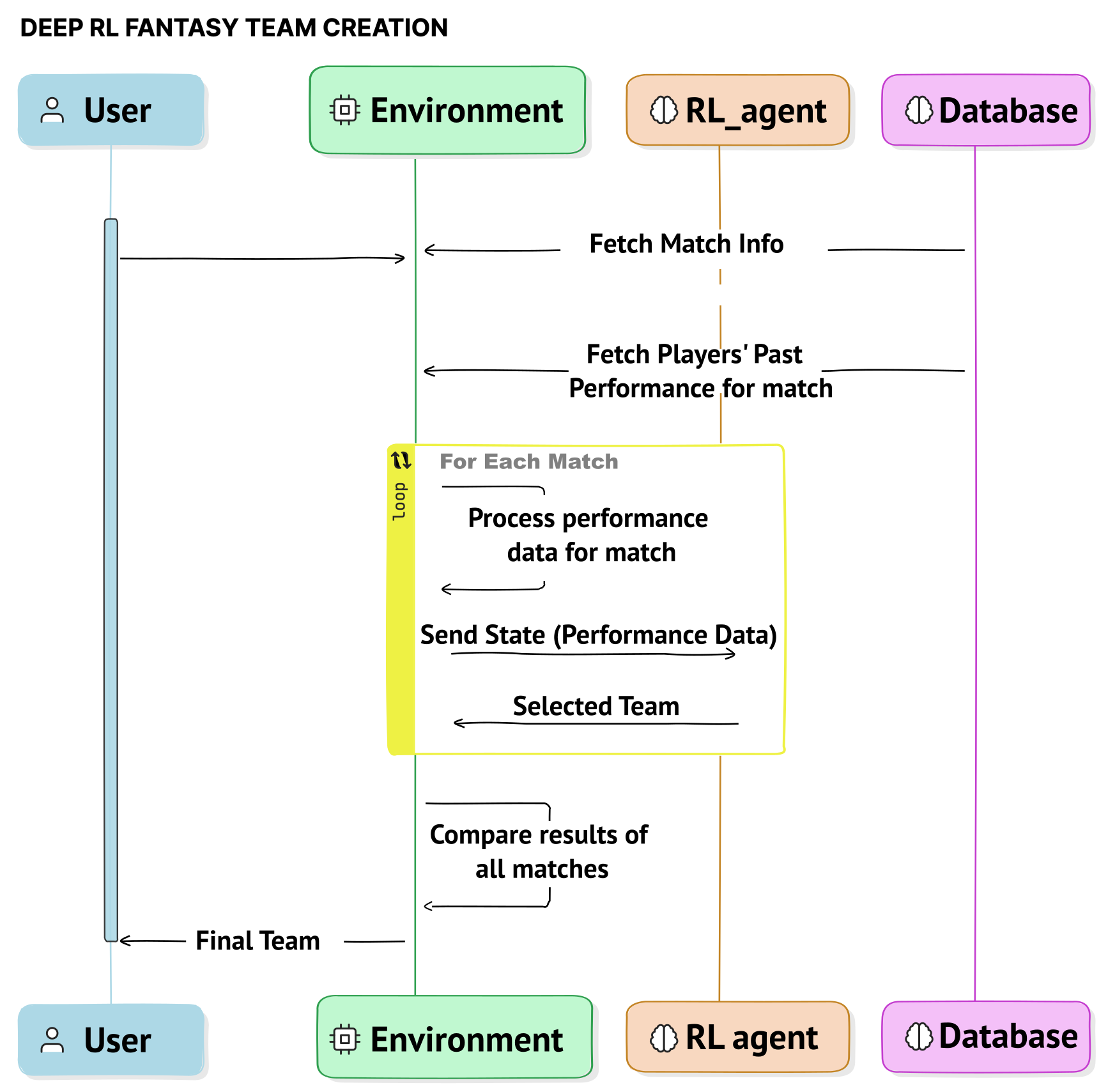}
    \caption{Our proposed fantasy team prediction system}
    \label{fig:proposed-model}
\end{figure}

To address this challenge, we created a reinforcement learning \cite{sutton2018reinforcement,silver2016mastering,mnih2015human,schulman2017proximal} based fantasy team creation agent to autonomously construct optimal cricket teams. By leveraging historical performance data, this research aims to provide a systematic framework for team composition, capable of adapting to changing match conditions and maximizing team performance. Key challenges to be addressed include the design of an RL framework capable of learning optimal team composition strategies, the development of effective state representations that capture the nuances of team dynamics and player performance, the definition of reward structures that incentivize desirable team compositions, and the evaluation of the effectiveness of the proposed methodology in enhancing team performance and competitiveness.

Through the exploration of these challenges and the development of a robust methodology, this research seeks to contribute to the advancement of data-driven decision-making in the fantasy sports industry. By providing our company users with valuable insights and systematic approaches to team composition, this research aims to enhance the overall competitiveness and success of cricket teams in fantasy contests.

In the next section, we will discuss some of the recent works related to our current research. This is followed by the Methodology section, where we detail the reinforcement learning algorithms employed, as well as the processes involved in data creation and pre-processing. Subsequent to this, we present our Experimental Results.

Once validated, the trained RL models can be deployed in real-world cricket scenarios to assist coaches and team selectors in making informed decisions regarding team composition. The RL-based approach offers a data-driven framework for optimizing team selection strategies, potentially enhancing team performance and competitiveness in professional cricket matches.

By leveraging RL techniques, this research aims to automate the process of team composition in cricket matches, offering teams valuable insights into crafting optimal lineups for improved performance. The proposed approach has the potential to revolutionize team selection strategies in cricket and other sports domains, paving the way for more data-driven decision-making and enhanced team competitiveness.

\section{Related Work}

In this section, we review the existing literature related to our work, focusing on three main areas: the evolution of fantasy sports, the application of machine learning in fantasy sports, and the use of reinforcement learning for team composition.

Fantasy sports have grown significantly since their inception in the 1960s with fantasy baseball. Over the years, the advent of the internet and advancements in technology have fueled the rapid growth and popularity of fantasy sports across various disciplines, including football, basketball, and cricket. Platforms like our company have capitalized on this growth, providing users with a sophisticated environment to create virtual teams and compete against each other.

When it comes to existing literature, although there has been some recent work on the application of machine learning for optimizing fantasy teams, the literature is scarce when it comes to reinforcement learning (RL) applications for this purpose. Machine learning techniques, such as regression analysis, decision trees, and neural networks, have been widely used to predict player performance and optimize fantasy sports teams. For instance, the review article \cite{bunker2022application} provided a comprehensive review of machine learning approaches used in predicting match results in team sport. In the paper \cite{beal2020optimising}, the authors used the players' past performance data using the linear regression, LSTM, radial basis functions (RBFs) and random forrest models. In the paper \cite{rajesh2022player}, the authors used ML techniques for making informed decisions in fantasy premiere league. In the article \cite{bonello2019multi} the authors incorporated external data sources other than just player performance features to create a better player performance prediction model.

Reinforcement learning (RL) has achieved remarkable success in tackling sequential decision-making problems across various fields. Examples include game playing \cite{silver2016mastering}, robotics control \cite{lillicrap2015continuous}, and resource management \cite{sutton2018reinforcement}. In these domains, RL agents learn optimal strategies through interaction with the environment and by receiving rewards for desired actions.
Combining fantasy sports with reinforcement learning techniques has been relatively unexplored and for that reason, the literature is extremely scarce. In the recent work, the authors \cite{macdow2018fantasy} proposed a deep RL-based approach for optimizing fantasy football team selection in Daily Fantasy Sports (DFS), demonstrating the potential of RL to outperform traditional methods. 

Our work builds on these foundations by applying RL to the problem of team composition in fantasy cricket. By leveraging historical performance data and RL algorithms like DQN and PPO, we aim to develop an autonomous agent capable of creating optimal cricket teams. Our research contributes to the growing body of knowledge in both fantasy sports and reinforcement learning, demonstrating the effectiveness of RL in enhancing team selection strategies and overall competitiveness in fantasy sports leagues. These recent advancements in RL, such as Proximal Policy Optimization (PPO) or Soft Actor-Critic (SAC), offer promising directions for further research in this area. These algorithms provide stable and efficient learning frameworks that can be leveraged to handle the complexities and uncertainties inherent in fantasy sports optimization.

\section{Methodology}

Historical performance data of individual players from both competing teams over the past ninety days is collected. This dataset includes metrics such as batting averages, bowling strike rates, fielding statistics, and other relevant performance indicators. The raw data is pre-processed to handle missing values, normalize numerical features, and encode categorical variables. This ensures uniformity and compatibility with the reinforcement learning models.

\subsection{Data Curation}
The datasets we used contain round-level player performance data. We focused on major T$20$ international matches, including the T$20$ World Cup, the Indian Premier League (IPL), and significant bilateral and trilateral T$20$ series. For any given round, there are a total of twenty-two playing members, eleven from each team. The data includes fantasy scores for each of the twenty-two players, with the eleven players having the most fantasy points constituting the dream team. The dream team may or may not be associated with a real user. Our task is to predict a team as close as possible to the dream team for each round.

Each round constitutes a single environment state. For each player, we consider the past ninety days of playing history, including the fantasy points scored. We then average these scores over the past ninety days to create a data point for each round. This results in a data shape of $(22, 10)$, where we use a total of ten features for each of the twenty-two players.

The training data comprises matches from January $2021$ to January $2023$, covering numerous high-stakes matches across various tournaments. For the test data, we use matches from March $2023$ to January $2024$, further splitting this period into validation and test sets. This temporal division ensures the evaluation of the model on data representative of real-world conditions, thereby enhancing its reliability and performance.

To address variability in scoring between different matches, we normalize the data at a match level. This normalization step ensures that the model treats high-scoring and low-scoring matches with equal importance, reducing bias towards higher-scoring matches and creating a balanced and fair representation of player performance across various matches.

\subsection{Reinforcement Learning Framework}

Reinforcement learning deals with optimal policy learning under uncertainty. The sequential decision-making problem is formulated as a Markov Decision Process (MDP). We collect data in the form of a tuple \( (S, A, R, S', A') \), where \( S \) represents the current state of the environment, \( A \) represents the action taken by the agent, \( R \) represents the reward received after taking action \( A \) in state \( S \), \( S' \) represents the new state of the environment after the action is taken and \( A' \) represents the action taken in the new state \( S' \).

\subsubsection{State Space}
At each decision point \( t \), the agent observes the current state \( O_t \in S \), which provides all the necessary information for making a decision at that specific moment. The state encapsulates recent performance data of all the players available in the current round. Specifically, the state \( O_t \) is expressed as \( O_t = (t, S_t, R_t) \), where \( t \) denotes the current time-step, \( S_t \) signifies the currently selected team, and \( R_t \) represents the players in reserve. The state space is thus defined by the composition of the current team, which includes both the selected eleven players \( S_t \) and the reserved players \( R_t \), with each player characterized by their performance metrics over the previous ninety days.

\subsubsection{Action Space}

The action space is composed of all possible player exchanges between the selected team and the reserve team. Let \( A \) represent the set of all possible actions, and \( A_t \) denote the action taken by the agent at an instant \( t \). Here, \( A_t \) is expressed as the tuple \( (rm_t, ad_t) \). In this tuple, \( rm_t \) refers to the player being removed from the selected team, while \( ad_t \) refers to the player being added to the selected team. Thus, each action corresponds to a player swap between the selected and reserve teams. The objective of the agent is to identify an optimal policy that allows it to choose an action (a player exchange) from the current state \( O_t \) that leads to an improved state (an optimal team). Starting with a randomly selected team, at each step, the RL agent will swap two players between the selected team \( S_t \) and the reserve team \( R_t \), thereby creating a new state \( O_{t+1} \). The agent will receive an immediate reward depending on the action \( A_t \) at the current state \( O_t \). The ultimate goal of the RL agent is to discover an optimal policy that maximizes the total cumulative reward. This action selection process is guided by the RL policy learned through either the DQN or PPO model.

\subsubsection{Reward Design}
At each instant $t$, the agent takes an action $A_t \in A$ from a state $O_t \in O$. In the process the agent gets an immediate reward $R_t$. The reward structure is crucial for guiding the agent's learning process. In our approach, the reward design is as follows:

\begin{itemize}
    \item \textbf{Step Reward:} At each step, the agent receives a reward of $-1$.  The $-1$ reward for each swap encourages the agent to minimize the number of swaps it takes to find the optimal team. This is necessary to prevent the agent from taking unnecessary swaps too much.
    \item \textbf{Soft Terminal State:} We introduce a soft terminal state at $\alpha \times$ the maximum possible score, where $\alpha$ is a hyperparameter chosen from the range $[0.7, 1.0]$. This soft terminal state helps in shaping the reward function and guiding the agent towards high-performing states even before reaching the actual goal state. More details about this hyperparameter is presented in the experimental evaluation section.
     
    \item \textbf{Goal State Reward:} When the agent reaches the goal state, it receives a substantial reward of $10$. This large positive reward is designed to reinforce the importance of reaching the goal state. The $+10$ reward value was chosen heuristically. This was to ensure that, during inference, the episodic reward for an ideal agent remains positive. In the worst-case scenario, the ideal agent would need to make eleven swaps to move from the least optimal team to the dream team. Therefore, by providing a +10 reward, the ideal agent will still maintain a positive cumulative reward after making the maximum number of necessary swaps. For any intermediate team, the ideal agent would require fewer than eleven swaps to reach the dream team, thus accumulating a positive reward.
\end{itemize}

The maximum possible selected team score represents the goal state, which is the desired outcome for the agent. There are no other terminal states in this setup, simplifying the reward structure and focusing the agent's learning towards achieving the goal state. This reward design encourages efficient exploration while motivating the agent to reach the goal state quickly while also providing guidance through the soft terminal state.

\subsubsection{Transition Function}
At every decision epoch $t$, the agent takes an action $A_t$ from the current state $O_t$ depending on the current policy $\pi_\theta(A_t | O_t)$. Due to this action by the agent $A_t$ at the current state $O_k$, the environment transitions to a new state depending on the state transition matrix $T(O_{t+1}|O_t,A_t)$. For our current problem setup, the transition function is deterministic, meaning that taking an action $A_t$ from a state $O_t$ will always change the environment state to a deterministic state $O_{t+1}$.

\begin{figure}[h]
\centering
\includegraphics[width=0.45\textwidth]{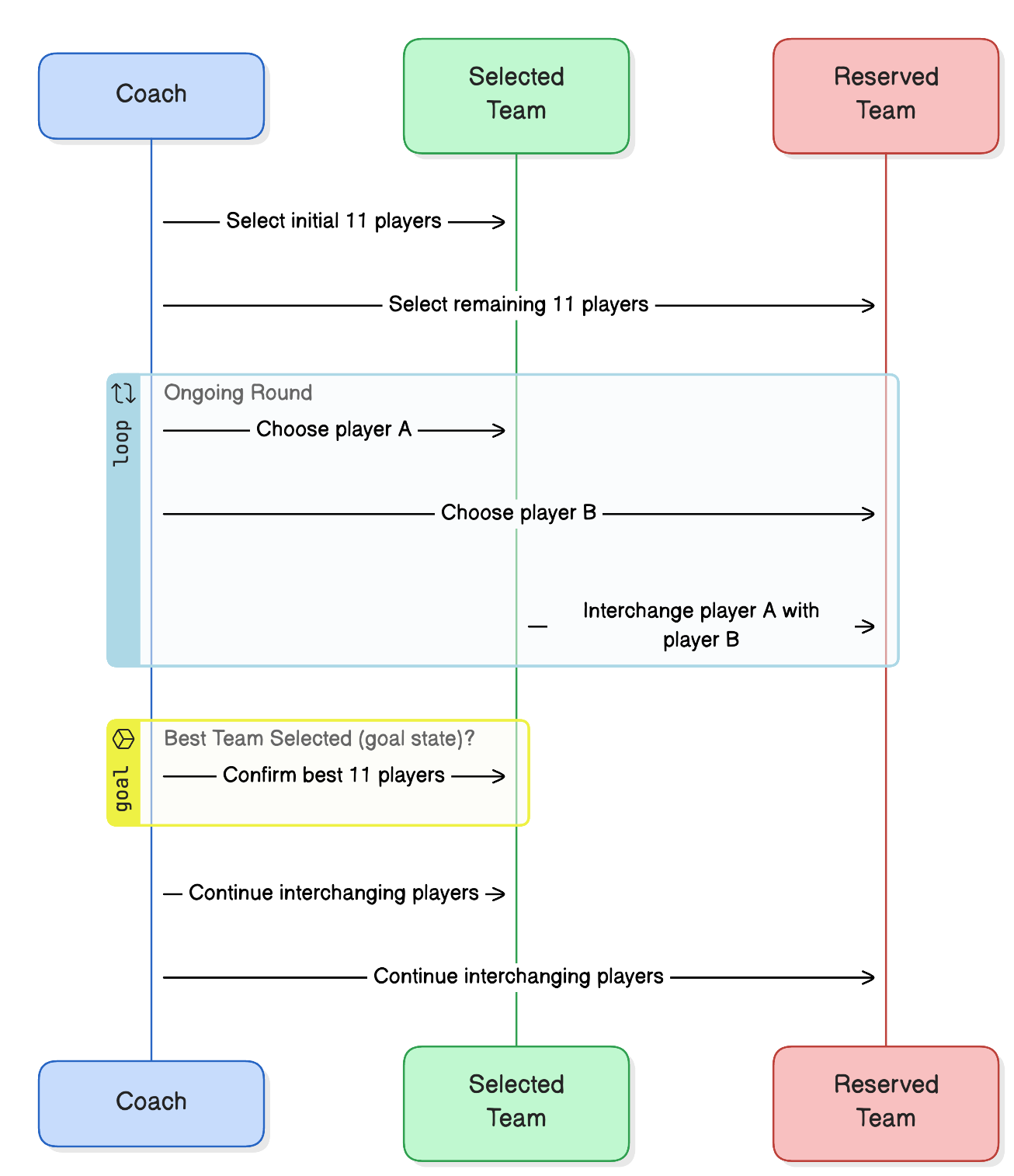}
\caption{The RL-agent environment interaction process}
\end{figure}

\subsubsection{Environment}
To facilitate the interaction between the reinforcement learning (RL) agent and the task at hand, we created a OpenAI Gym \cite{brockman2016openai} compatible environment. This custom environment enables the agent to generate experience points by interacting with it iteratively. At the start of each episode, the environment returns a randomly selected team to the agent. Based on the currently selected state, the agent decides on an action to modify the team. This iterative process continues until the goal state criteria are met, indicating that the desired team configuration has been achieved. Upon reaching the goal state, the environment resets and provides a new randomly selected team for the agent to engage with in a subsequent episode.

\subsection{RL architectures}
Two distinct reinforcement learning algorithms, namely Deep Q-Networks (DQN) \cite{mnih2013playing} and Proximal Policy Optimization (PPO) \cite{schulman2017proximal}, are employed to tackle the team composition problem.

\subsubsection{Deep Q-Networks (DQN)}
DQN utilizes a Q-learning approach to estimate the optimal action-value function \(Q(s, a)\), where \(s\) represents the state (current team composition) and \(a\) represents the action (player swap). The Q-network is trained to minimize the temporal difference error \cite{sutton2018reinforcement} between predicted and target Q-values:

\[L_{DQN} = \left(r + \gamma \max_{a'} Q(s', a') - Q(s, a)\right)^2\]

where \(L_{DQN}\) is the loss function for DQN, \(r\) is the reward, \(\gamma\) is the discount factor, and \(s'\) is the next state.

\subsubsection{Policy-Based Methods}
Policy-based methods \cite{sutton1999policy,sutton2018reinforcement,schulman2017proximal} optimize the policy \(\pi(a|s)\) directly, aiming to find the best action distribution given a state. The objective is to maximize the expected return by updating the policy parameters \(\theta\) to increase the probability of taking better actions:

\[J(\theta) = \mathbb{E}_{s \sim d^\pi, a \sim \pi_\theta} \left[ \sum_{t=0}^\infty \gamma^t r(s_t, a_t) \right]\]

where \(J(\theta)\) is the objective function for policy-based methods, \(d^\pi\) represents the state distribution under policy \(\pi\), and \(\gamma\) is the discount factor.

In this work, we use a specific policy-based algorithm called Proximal Policy Optimization (PPO) \cite{schulman2017proximal}. PPO optimizes the policy function while ensuring the policy update stays within a specified range to avoid drastic changes, which is achieved through a clipped surrogate objective.

The neural network architecture used in our PPO model consists of a custom feature extractor and a policy network. The feature extractor first flattens the input observations and then processes them through three fully connected layers with 256, 512, and 1024 units, respectively, each followed by a Tanh activation. This generates a rich latent feature representation. The policy network further processes these features through two fully connected layers with 256 and 512 units, respectively, again using Tanh activations. The final layer produces logits corresponding to the action space size, which are converted into action probabilities using a softmax function. This architecture efficiently maps high-dimensional observations to action probabilities, ensuring smooth training and effective policy learning.

Both the RL algorithms use neural network architectures to optimize the policy for our task, and these architectures are crucial for determining the agent's performance. The $PPO$ model utilizes a single neural network with two output heads: one for the policy (actor) and one for the value function (critic). This shared network is responsible for both selecting actions and evaluating the value of the current state. The shared network acts as the feature extraction layer. This layer processes input data through three fully connected layers with $256$, $512$, and $1024$ units, respectively, each followed by a ReLU activation function. The policy network further processes these features through two fully connected layers with 512 and 256 units, respectively, each followed by ReLU activation function. The final layer of the policy head produces logits corresponding to the action space size, which are converted into action probabilities using a softmax function. The value network uses the same feature extractor network, after which it uses a single dense layer to project the features to a single output value. This output layer of the value function network uses a linear activation function.

The DQN model uses a separate neural network to approximate the $Q$-values for each action in a given state. The architecture for the DQN model includes an input layer with neurons corresponding to the state representation, followed by two hidden layers with $256$ neurons each, using ReLU activation functions. The output layer produces $Q$-values for each action in the action space using a linear activation function.

For both the PPO and DQN models, we employ the Adam optimizer for training, with a learning rate of \(1 \times 10^{-3}\). A batch size of $128$ is used to ensure computational efficiency and training stability. The networks are trained over $10000$ episodes, with each episode consisting of a variable number of time steps, depending on task complexity. A discount factor (\(\gamma\)) of $0.99$ is used to balance immediate and future rewards. Table \ref{tab:training_settings} details all the key settings used for training both the DQN and PPO models.

\begin{table}[h]
    \centering
    \caption{Training setup for DQN and PPO Models}
    \label{tab:training_settings}
    \begin{tabular}{lcc}
        \toprule
        \textbf{Parameter} & \textbf{DQN} & \textbf{PPO} \\
        \midrule
        Number of Time-steps & 2,000,000 & 2,000,000 \\
        Learning Rate & 0.0001 & 0.0001 \\
        Gamma ($\gamma$) & 0.99 & 0.99 \\
        Batch Size & 128 & 128 \\
        Replay Buffer Size & 10,000 & - \\
        Target Network Update Frequency & 5,000 steps & - \\
        Exploration Fraction & 0.1 & - \\
        Exploration Final $\epsilon$ & 0.02 & - \\
        Number of Environments & - & 8 \\
        Number of Epochs & - & 10 \\
        Clip Range & - & 0.2 \\
        Value Function Coefficient & - & 0.5 \\
        Entropy Coefficient & - & 0.01 \\
        \bottomrule
    \end{tabular}
\end{table}

\begin{figure}[h]
    \centering
    \includegraphics[width=0.5\textwidth]{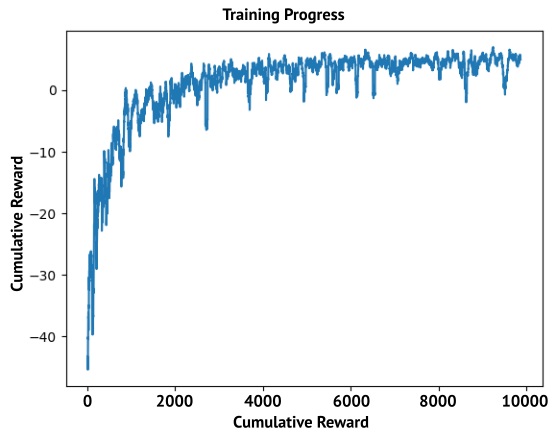}
\caption{Training Progress of PPO Model. The plot shows the cumulative reward over 10,000 episodes of training. The cumulative reward initially starts at a negative value and gradually increases as the training progresses. Eventually, the reward stabilizes, demonstrating the convergence of the PPO model towards optimal behavior.}
\label{fig:ppo-training}
\end{figure}


\begin{figure}[h]
    \begin{center}
            \includegraphics[width=0.5\textwidth]{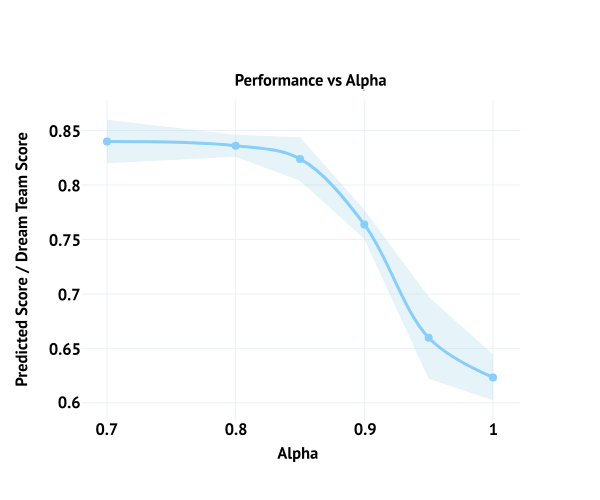}
        \caption{This plot illustrates the relationship between the alpha values and the corresponding performance of the model. The shaded area represents the error bounds for each alpha value obtained using the cross-validation datasets}
        \label{fig:alpha}
    \end{center}
\end{figure}

\begin{figure}[h]
    \begin{center}
            \includegraphics[width=0.5\textwidth]{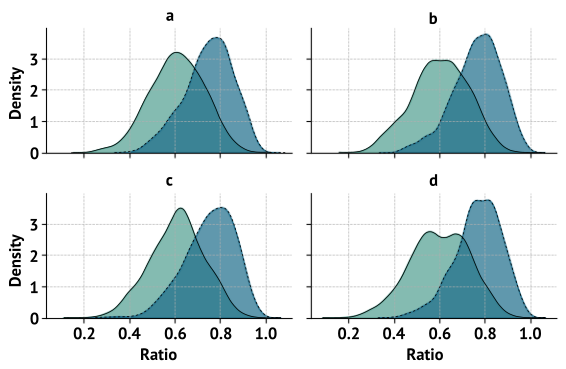}
        \caption{Density plots showing the distribution of the ratio between the predicted team score and the best team score before and after training across four different cross-validation datasets (a, b, c, and d). In each plot, the solid line represents the distribution of the ratio before training, while the dotted line represents the distribution of the ratio after training. Each plot illustrates how the density shifts towards higher ratios post-training, indicating an improved performance of the agent in selecting teams with scores closer to the best possible team score}
        \label{fig:ratio_density_plots}
    \end{center}
\end{figure}

\begin{table}[ht]
\centering
\caption{Percentile scores comparing fantasy teams predicted by the trained RL agent with those created by other common team creation strategies, relative to real user teams}
\label{tab:percentile_scores}
\begin{tabular}{lcccc}
\toprule
 & \multicolumn{4}{c}{Test set result} \\
\cmidrule(r){2-5}
 & (cv1) & (cv2) & (cv3) & (cv4) \\
\midrule
Previous Performance & 0.54 & 0.51 & 0.57 & 0.54 \\
Player \% selection & 0.55 & 0.56 & 0.54 & 0.51 \\
\midrule
RF classifier & 0.57 & 0.54 & 0.51 & 0.56 \\
SVM classifier & 0.55 & 0.56 & 0.54 & 0.55 \\
\midrule
DQN agent & 0.61 & 0.58 & 0.59 & 0.52 \\
PPO agent & 0.67 & 0.62 & 0.64 & 0.62 \\
\bottomrule
\end{tabular}
\end{table}

\section{Experimental Evaluation}

\begin{algorithm}
\caption{PPO for Fantasy Team Creation}\label{alg:ppo_fantasy}
\begin{algorithmic}[1]
\State \textbf{Input:} Players: List of $22$ players with historical scores
\State \textbf{Input:} SelectedTeam: Randomly selected team of $11$ players
\State \textbf{Input:} ReservedTeam: Remaining $11$ players
\State \textbf{Input:} $T$: Maximum number of steps per episode
\State \textbf{Input:} $\theta$: Parameters of RL agent

\State Initialize parameters $\theta$ of RL agent
\For{episode = 1, 2, \dots}
    \State $\tau \gets \emptyset$
    \For{$t = 0, 1, \dots, T-1$}
        \State Select a player $p_{\text{sel}}$ from SelectedTeam
        \State Select a player $p_{\text{res}}$ from ReservedTeam
        \State Interchange $p_{\text{sel}}$ and $p_{\text{res}}$ between SelectedTeam and ReservedTeam
        \State Calculate reward $r_t$
        \If{SelectedTeam has high historical scores}
            \State $r_t \gets 10$
        \Else
            \State $r_t \gets -1$
        \EndIf
        \State Add \{$\text{SelectedTeam}, p_{\text{sel}}, p_{\text{res}}, r_t$\} to $\tau$
    \EndFor
    \State Update model parameters $\theta$
\EndFor
\State \textbf{return} trained policy
\end{algorithmic}
\end{algorithm}

Algorithm \ref{alg:ppo_fantasy} outlines the procedure for training our RL agents to create a fantasy cricket team. The algorithm starts by preparing a list of $22$ players, which are divided into two groups: the initially selected team of $11$ players and the reserved team of the remaining $11$ players. During each episode, the agent recursively swaps players between the selected and reserved teams. The agent receives a reward of $-1$ for each step to encourage efficient convergence and a reward of $+10$ when it reaches a team configuration with high historical scores, defined as the goal state. This process is repeated over multiple episodes to optimize the agent's policy.

We use the hyperparameter $\alpha$ to control how close a predicted team's score must be to the best team's score in order to be classified as a goal state. When $\alpha$ is close to 1, the agent is required to predict teams that perform almost as well as the best possible team, making it difficult to reach the goal state. This leads to slower training progress and potentially lower accuracy after a fixed number of training timesteps. However, teams close to the dream team score should also be considered goal states, which effectively reduces noise and creates more goal states for the agent to learn from. By lowering $\alpha$, we make it easier for the agent to receive rewards by accepting teams that perform $\alpha$ times the score of the best team. This results in the agent more frequently reaching the goal state, leading to an improvement in cumulative episodic reward and enabling faster learning with improved accuracy within a limited training duration. However, lowering $\alpha$ too much allows suboptimal teams to be classified as goal states, which can reduce the quality of the teams selected by the agent. Thus, there is a trade-off between cumulative reward and team selection accuracy. To study this behavior, we observed the accuracy of team selections at fixed training intervals for different values of $\alpha$ (see Fig. \ref{fig:alpha}). Our goal is to select an $\alpha$ as close as possible to 1 without degrading the RL agent's training performance, and accordingly, we selected 0.8 as the value for the hyperparameter $\alpha$.

In this study, we used the Stable-Baselines3 \cite{raffin2021stable} library to create and train our reinforcement learning (RL) agents. The computational tasks were executed on the Databricks platform, employing multiple GPUs to facilitate distributed training. A four-fold cross-validation approach was adopted to train and validate the models. We ensure that both the training and validation sets respect the temporal ordering of the data. In each fold of cross-validation, the validation data consists of rounds that occur after the rounds in the training data. A suitable temporal gap between the training and validation data is maintained. This is done to ensure that no future data is used to train the model, thereby avoiding data leakage. For example, in a given validation-fold, if rounds from timesteps $t_0$ to $t_4$ are used for training, then rounds from $t_5$ to $t_6$ are used for validation. A suitable gap (timesteps $t_4$ to $t_5$) is maintained between the training and validation rounds. In the next fold, rounds $t_0$ to $t_5$ are used for training, and rounds from $t_6$ to $t_7$ are used for validation, and so on.

Figure \ref{fig:ppo-training} shows the cumulative reward received by the Proximal Policy Optimization (PPO) agent across $10000$ episodes during training. The y-axis represents the cumulative reward, while the x-axis represents the number of episodes. The cumulative reward is low initially and as training progresses, the cumulative reward steadily increases and stabilizes, demonstrating that the agent is effectively learning to maximize the reward. The fluctuations in the reward during training indicates the variability in the quality of the initial teams.

To validate the performance of the trained agents, we compared the agent-predicted teams with baseline machine learning models, specifically Support Vector Machine (SVM) and Random Forest (RF) classifiers. The SVM classifier was trained using an RBF kernel with the default regularization parameter value of $C$ = 1. The Random Forest model was trained using 100 trees with a maximum depth of 10. For both baseline models, hyperparameters were chosen using the grid-search method on cross-validation datasets. We also compared the performance of the RL agent-predicted teams with other popular team creation strategies. These strategies included creating teams based on the best-performing players from the last match and selecting players with the highest selection percentage by fantasy users in the current round (at the time of prediction, before round-lock). The comparative results are detailed in Table \ref{tab
}. The table shows a comparison of team performance using traditional team creation methods (Previous Performance and Player 

The comparison reveals that while the SVM and Random Forest models generally outperform traditional methods like Previous Performance and Player \% Selection, their performance is still inferior to the RL agents. In general, users in the top 40th percentile win some rewards in the game. Notably, the teams generated by our RL agents consistently placed above the 60th percentile, on average, suggesting that users have a higher likelihood of winning if they utilize the RL agent for team creation in any round. The results also indicate that the PPO agent consistently outperforms both the baseline models and the DQN agent, demonstrating its effectiveness in selecting higher-performing teams.

This suggests that while machine learning models can capture some aspects of optimal team selection, reinforcement learning agents are better equipped to handle the complexity of sequential decision-making and long-term planning. The PPO agent consistently achieved the highest scores across all cross-validation sets, followed closely by the DQN agent, demonstrating the advantage of using RL for dynamic team selection.

To further validate the model performance, we checked the predicted team scores wrt the actual dream team scores for any given test round(Figure \ref{fig:ratio_density_plots}). The figure shows the density plots for the ratio of the predicted team score to the actual dream team score for the corresponding round. The different plots (a, b, c, and d) in the figure represents four different cross-validation datasets. Each plot in the figure show the distribution of the ratio generated by a random agent (green-shaded curve) and by the trained agent (blue-shaded curve). Each plot illustrates how the density shifts towards higher ratios post-training, indicating an improved performance of the agent in selecting teams with scores closer to the best possible team score.

\section{Conclusions \& Discussions}
In this paper, we presented a novel approach to fantasy team creation using deep reinforcement learning, specifically focusing on the application of the Deep Q-Learning (DQN) and the Proximal Policy Optimization (PPO) agents. Our method effectively leverages historical player performance data to create teams with a higher likelihood of winning. Through extensive experimentation, our results indicate a significant shift in the distribution of team scores towards higher percentiles post-training, demonstrating an increased likelihood of the teams selected by our model ending within the \textbf{winning zone} (above the 60th percentile). This trend is consistently observed across multiple cross-validation datasets, suggesting that the agent's training has effectively improved its ability to select higher-performing teams.

Our model's performance was benchmarked against traditional team creation techniques, which are widely used in the fantasy sports community. Our model have shown significant improvement over these traditional team creation techniques. This improvement can be attributed to the agent’s ability to recognize patterns in player performance and make unbiased team selections. Unlike human users who might have inherent biases towards or against certain players, ML models can evaluates players purely based on their potential to maximize the team's success. This not only removes the emotional component of team selection but also enhances the overall quality of the team.

One of the key implications of our work is its potential to engage new users in fantasy sports. By providing new users with a strong baseline team, they can have an improved gaming experience, which could enhance user retention and satisfaction. Additionally, our model serves as a useful tool for less experienced users, providing them with a competitive edge in fantasy leagues. While this paper has focused primarily on cricket, the underlying principles of our method have broader applicability across various fantasy sports. This flexibility makes our model a versatile tool for the fantasy sports domain.

Future work could explore several avenues for enhancing the model’s performance further. For instance, integrating real-time player performance data could enable dynamic team adjustments, allowing for even more responsive and accurate team selections. Additionally, exploring other reinforcement learning techniques or hybrid models could potentially yield even better results.

\bibliographystyle{ACM-Reference-Format}
\bibliography{manuscript}

\end{document}